\documentclass[runningheads]{llncs}
\usepackage[T1]{fontenc}
\usepackage{graphicx}
\usepackage{booktabs}
\usepackage[misc]{ifsym}
\usepackage{custom-packages}

\usepackage{hyperref}
\usepackage{multicol}
\usepackage{threeparttable}

\begin{document}

\title{
Improving Wind and Solar Power Prediction with Efficient Wrapper-based Feature Selection:\\An Empirical Study
}
\titlerunning{Performance-Driven Feature Selection for Renewable Energy Prediction}
%
\author{Daniel Grillmeyer\inst{1}\orcidID{0009-0009-0997-4227} \and
Marius Hadry\inst{1}\orcidID{0000-0002-3602-5790} \and
Michael Stenger\inst{1}\orcidID{0009-0009-5850-2445} \and
Vanessa Borst\inst{1}\orcidID{0009-0004-7123-7934} \and
Veronika Lesch\inst{2}\orcidID{0000-0001-7481-4099} \and
Samuel Kounev\inst{1}\orcidID{0000-0001-9742-2063}}

\authorrunning{Grillmeyer et al.}

\institute{
University of Würzburg, Würzburg, Germany\\
\email{\{Daniel.Grillmeyer,Marius.Hadry,Michael.Stenger,Vanessa.Borst,\\Samuel.Kounev\}@uni-wuerzburg.de} \and
Baden-Wuerttemberg Cooperative State University Mosbach, Mosbach, Germany\\
\email{Veronika.Lesch@dhbw.de}}

\maketitle              

\begin{abstract}
With rising global energy demand and growing awareness of climate change and its impacts, the share of renewable energies in the global energy mix continues to grow.
Unlike conventional power generation, the output of renewable energy sources cannot be controlled as consistently due to their dependence on environmental conditions. 
Therefore, reliable prediction of current and future energy production is essential.
In this paper, we report findings from two structured literature reviews on real-world renewable energy prediction tasks: wind turbine power curve modeling and photovoltaic power prediction. For the former, we conducted a comprehensive literature review ourselves, while for the latter, we synthesize the key findings regarding frequently selected input features based on an existing survey. Across both domains, our analysis reveals that despite the large number of available monitoring and environmental variables, only limited or unsystematic methods for feature selection exist.
To address this gap, we propose Cluster-based Sequential Feature Selection (CSFS), a novel, model-agnostic, clustering-based wrapper method for automatic, efficient, and reliable feature selection in renewable energy prediction pipelines. To support reproducibility and reuse, we provide an open-source implementation of CSFS on \href{https://github.com/dan-grm/CSFS-preprint}{GitHub}.
We empirically evaluate the proposed approach on both use cases and compare it with established feature selection techniques such as wrapper-based sequential feature selection (SFS), filter-based methods, and Random Forest's embedded feature importance.
The results show that the wrapper-based methods overall provide better-performing selections of features. CSFS achieves a predictive performance comparable to SFS while reducing computational cost by an average of 21\%.

\end{abstract}

\section{Introduction}

The global energy demand has constantly increased in the last two decades from 430 Exajoules to 590 Exajoules~\cite{PARASCHIV2023276}. 
In the era of climate change, the generation and use of renewable energies is gaining in importance.
Accordingly, global power production from renewable energies has more than doubled in the last ten years, from 1,851,114~MW to 4,448,051~MW in 2024 worldwide~\cite{Irena25_Energy} with wind and solar being the fastest-growing areas increasing by 16\% in 2024~\cite{PARASCHIV2023276}. 
To ensure a reliable power supply, energy suppliers must accurately estimate the amount of power generated.
While electricity production from conventional sources such as fossil fuels and nuclear energy can be adjusted relatively precisely, renewable energy sources are often dependent on environmental factors~\cite{PARASCHIV2023276}.

To enable accurate power estimation despite strong environmental dependencies, the literature proposes numerous approaches for power prediction in wind and solar energy systems. An existing literature review on \ac{PV} power prediction--- analyzing 112 articles~\cite{gupta_pv_2021}---together with our own comprehensive literature study on \ac{WTPC} modeling, comprising 218 publications, 
shows that a diverse set of input features can be used. These include, among others, atmospheric pressure, humidity, wind direction, ambient temperature, and solar irradiance. 
However, while some approaches use \ac{FS} methods and multiple features, most approaches only use a small subset of features that hardly differ from each other. 
Therefore, one could assume that there is unexploited potential in the features not considered~\cite{machines9050100}, which can only be determined through a comprehensive feature selection procedure.
There exist filter-based, wrapper-based, and embedded methods for \ac{FS}. While filter methods are considered most efficient, they have drawbacks in predictive performance with the resulting feature set.
Wrapper methods are computationally more expensive, but are capable of finding a better solution.
Embedded methods are typically faster than wrapper methods, but do not offer a model-agnostic adoption, and might not find an optimal feature set~\cite{DBLP:journals/apin/DhalA22}.

In this paper, we propose a novel clustering-based approach that aims to balance the low computational cost of classical filter-based \ac{FS} methods with a more comprehensive assessment of feature relevance for predictive performance. More specifically, our contributions are threefold:

 \textbf{1. Insights into frequently used features and modeling techniques for wind and solar energy:} We conduct an extensive literature analysis on wind turbine power forecasting, analyzing 218 papers to derive insights into commonly used input features and modeling techniques. In addition, we synthesize the findings of an existing review on photovoltaic power generation~\cite{gupta_pv_2021} to obtain comparable insights for photovoltaic power prediction. To ensure transparency and reproducibility, we provide the complete code, the results underlying our summaries of both reviews, and the analysis scripts, on \href{https://github.com/dan-grm/CSFS-preprint}{GitHub}.

\textbf{2. Cluster-based Sequential Feature Selection (CSFS)}: We propose CSFS, a novel, model-agnostic, and cluster-based methodology for automated \ac{FS}. The core idea is to apply a clustering technique to first identify feature relevance at a coarse-grained level before analyzing features at a fine-grained level. This two-stage procedure aims to reduce complexity while still ensuring that all relevant features are considered. In principle, the proposed methodology is applicable to a wide range of \ac{FS} problems. To facilitate reuse and further development, we provide a scikit-learn~\cite{scikit-learn}-compatible open-source implementation on \href{https://github.com/dan-grm/CSFS-preprint}{GitHub}. By making the code publicly available, we aim to support reproducibility and transparency, and encourage the \ac{ML} community to adopt, scrutinize, and further extend the proposed approach.
    
\textbf{3. Empirical study based on two representative use cases}: 
We evaluate the proposed method in two application domains, namely \ac{WTPC} modeling and \ac{PV} power prediction, and perform a detailed comparison between our wrapper-based \ac{FS} approach and established techniques. Our results overall confirm that wrapper-based methods are capable to find better-performing feature sets than simpler baseline methods, with \ac{RMSE} values of $(32.49 \pm 0.30)$~kW for a 2~MW \ac{WT} and $(0.971 \pm 0.047$)~MW for a 20~MW \ac{PV} power station, respectively.
At the same time, the proposed method \ac{CSFS} reduces computational runtime compared to the established wrapper-based approach SFS, achieving an average reduction of 21\%, while achieving comparable predictive performance. These results suggest that CSFS provides a reasonable alternative that enables faster time-to-result while delivering reasonable results.

The remainder of the paper is structured as follows.
First, Section~\ref{sec:rw} introduces \ac{WT} and \ac{PV} power generation, summarizes the most frequently used features and modeling techniques in these domains based on two literature reviews, and further discusses relevant related work. 
Section~\ref{sec:approach} states the addressed problem definition, and presents our novel approach on model-agnostic cluster-based sequential \ac{FS}.
Section~\ref{sec:experimental_setup} provides information on the experimental setup, 
while Section~\ref{sec:results} analyzes and discusses the results of the experiments.
Finally, Section~\ref{sec:conclusion} summarizes the findings of the paper and highlights future work.
\section{Literature Review and Related Work}
\label{sec:rw}


\subsection{Wind Turbine Power Prediction}
\label{sec:rw:wind}

In a \ac{WT}, the kinetic energy of moving air is first transferred to the shaft in the form of mechanical energy and then converted into electrical energy by the generator. Since a \ac{WT} cannot reduce the wind speed to zero, it is physically impossible to extract all kinetic energy from the airflow. The actual efficiency with which the turbine converts the kinetic energy of the wind into mechanical power is therefore characterized by the power coefficient $C_p$. 
The actual power produced by a wind turbine can be expressed as $P_{\text{WT}} = 0.5 C_p A \rho v^{3}$, where $A$ is the rotor swept area, $\rho$ represents the air density, and $v$ is the wind speed. 
The theoretical maximum value of $C_p$ is defined by the Betz limit, which states that no wind turbine can extract more than 59.3\% of the kinetic energy contained in the wind stream:
$C_{p,\text{Betz}} = \frac{16}{27} \approx 0.593$. Modern \acp{WT} typically operate below this theoretical limit, achieving power coefficients of approximately $C_p \approx 0.5$~\cite{hansen_aerodynamics_2008}.

A \acf{WTPC} describes the relationship between the wind speed~$v$ and power~$P_{\text{WT}}$. 
Due to a lack of comprehensive literature reviews on \ac{WTPC} input features, we perform a \ac{SLR} to identify the used modeling approaches and input features.
For the \ac{SLR}, we follow the berry picking information retrieval model~\cite{bates_design_1989}, which iteratively refines search queries as new insights emerge, starting with the initial query ``wind power curve'' and considering the first 30~results on Google Scholar from 11~different queries. Due to overlaps, we identified 218 publications. To focus on the most recent papers, we discarded 90 results from 2014 and older, and further 38~results due to text unavailability or non-relatedness to \ac{WTPC} modeling, yielding a final body of 90~papers.
%
%
The literature on \ac{WTPC} modeling encompasses a variety of methodological paradigms, including statistical, kernel-based, and Bayesian models, as well as generalized linear and neural network approaches. 

\begin{figure}
    \centering
    \includegraphics[width=1\linewidth]{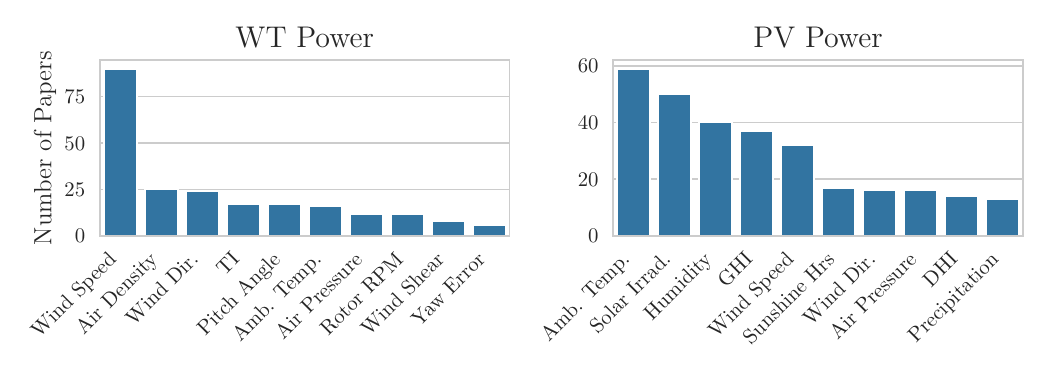}
    \caption{Top ten most frequently used features in \ac{WT} and \ac{PV} power prediction.}
    \label{fig:rw:feature_count}
\end{figure}

Besides wind speed, which is the physically most impactful explanatory variable, prior work has identified relationships with additional environmental features, such as air pressure, humidity, and wind direction, as well as operational features, such as blade pitch angle, component temperatures, and rotor speed. Fig.~\ref{fig:rw:feature_count} shows the frequency of the ten most frequently used input features. Wind speed is used as an input variable in all 90~papers, whereas the frequency of the subsequent features decreases significantly. For instance, air density and wind direction are used in only 25 and 24 papers, respectively. For the remaining features, the frequency further decreases to only 6 papers for the yaw error. 
Beyond the results shown in Fig.~\ref{fig:rw:feature_count}, 38 papers rely solely on wind speed, and the vast majority of 85.6\% of papers employ no more than five features for \ac{WTPC} modeling. Only 4.4\% of the reviewed papers provide the source code, and for only 17.8\%, the dataset is provided or available on request. \Ac{RMSE}, \ac{MAE}, and $R_2$-Score are the three most established performance metrics with 64\%, 53\% and 30\% usage ratio, respectively.
The complete results of our literature review, including a comprehensive CSV file containing all retrieved and analyzed publications as well as the evaluation scripts, are available on \href{https://github.com/dan-grm/CSFS-preprint}{GitHub}.




\subsection{Photovoltaic Power Prediction}
\label{sec:rw:pv}
The generation of power in a \ac{PV} cell is based on the photovoltaic effect. Its discovery is attributed to Alexandre-Edmond Becquerel in 1839, who experimented with metal electrodes in acid, neutral, and alkaline solutions and demonstrated a voltage when the two electrodes were unevenly exposed to sunlight~\cite{sharma_comprehensive_2025}. \ac{PV} cells consist of semiconductor sheets that convert solar energy into electrical energy, because sun exposure leads to an electric field between the layers, and therefore an electric flow~\cite{tyagi_progress_2013}.

\Ac{PV} power prediction aims to estimate the power output of a solar plant. Gupta and Singh~\cite{gupta_pv_2021} show in a comprehensive study that supervised learning is widely used. In their work, they emphasize that ensemble learning-based models show an advantage over single-stage forecasting methods such as \ac{SVM} or \ac{KNN}, but that deep learning methods have also become established in recent years.


As shown in Fig.~\ref{fig:rw:feature_count}, existing studies on \ac{PV} power forecasting employ a broad range of atmospheric and meteorological input features, including wind speed and direction, ambient temperature, humidity, atmospheric pressure, and various measures of solar irradiance~\cite{basaran_systematic_2020,gupta_pv_2021}. 
Additional variables derived from solar geometry, such as solar altitude, zenith, elevation, and the equation of time, can be computed from site coordinates, pressure, and temperature. 
Some approaches also incorporate clear-sky estimates of the \ac{GHI}, \ac{DNI}, and \ac{DHI}~\cite{gupta_pv_2021}, which can be calculated from solar position variables following the models proposed by Ineichen and Perez~et~al.~\cite{ineichen2002new,perez2002new}. 

Despite this wide variety of physically meaningful and potentially informative predictors, to the best of our knowledge no prior work systematically evaluates all of these predictors in a rigorous feature selection study. This observation is consistent with findings from previous surveys~\cite{basaran_systematic_2020,gupta_pv_2021}.
The results of our re-evaluation of the survey by Gupta and Singh~\cite{gupta_pv_2021}, including the extracted feature statistics and the analysis scripts used to generate Fig.~\ref{fig:rw:feature_count}, are provided at \href{https://github.com/dan-grm/CSFS-preprint}{GitHub}.

\subsection{Overview of Feature Selection Techniques}
\label{sec:rw:feature-selection}

\Ac{FS} methods are used to reduce the dimensionality of a dataset. Selecting only the most relevant features can increase the accuracy of \ac{ML} models, while decreasing the computational complexity. \Ac{FS} techniques can be categorized into filter, wrapper, and embedded methods~\cite{DBLP:journals/apin/DhalA22}.

Filter methods operate independently of the \ac{ML} model intended for prediction. They rather select a set of features that meet pre-defined criteria based on a feature relevance measure. Typical scoring functions for regression tasks are the F-value or \ac{MI} to measure the dependency between a feature and the target variable. Filter methods are considered most efficient but have drawbacks in predictive performance for the resulting feature set~\cite{DBLP:journals/apin/DhalA22}. 

Wrapper methods use the \ac{ML} model to assess the suitability of specific feature sets using evaluation metrics. They are computationally more expensive than filter methods, since they require fitting the \ac{ML} model to iteratively assess feature subsets, but are capable of finding a better solution~\cite{DBLP:journals/apin/DhalA22}.

Embedded methods leverage model-specific properties of \ac{ML} models to obtain a ranking of feature importance, for example, using impurity measures for tree-based algorithms. They offer a trade-off as they are typically faster than wrapper methods, but might not find an optimal feature subset, for instance, due to a bias towards high-cardinality features. Further, only selected \ac{ML} algorithms offer an intrinsic feature importance measure, limiting the applicability.

\section{Approach}
\label{sec:approach}

\subsection{Problem Definition}
\label{sec:approach:problem_definition}

The problem addressed in this work is a bi-level optimization problem that aims to jointly optimize feature set selection and model parameters in data-intensive ML-driven information systems.


Each dataset $\mathcal{D}=(\bm{X}, \bm{Y})$ contains a temporally ordered sequence of observations $\bm{x}_i \in \mathbb{R}^{|\mathcal{F}|}$, where $\mathcal{F}$ is the set of all available input features. The target variables $\bm{y}_{i} \in \mathbb{R}$, in our use cases, represent the power output at the $i$-th timestamp.
%
%
%
%
%
We define $\mathcal{D_{\mathcal{F'}}}=(\bm{X}_{\mathcal{F'}}, \bm{Y})$ as a feature-reduced version of dataset $\mathcal{D}$ with feature vectors $\bm{x}_i \in \mathbb{R}^{|\mathcal{F'}|}$ and $\mathcal{F'} \subseteq \mathcal{F}$. Further, we denote the set of feature subsets of size $n$ as $\mathcal{S}_n = \{\, \mathcal{F'} \subseteq \mathcal{F} \mid |\mathcal{F'}| = n \,\}$.

The bi-level optimization problem can be described as

\begin{equation}
    \mathcal{F}_{n}^{*}
= \arg\min_{\mathcal{F'} \in \mathcal{S}_n}
\left[
\min_{\theta \in \Theta,\; \lambda \in \Lambda}
\mathcal{L}\!\left(
f(\bm{X}_{\mathcal{F'}};\theta,\lambda),
\bm{Y}
\right)
\right],
\label{eq:approach:problem_definition:optimization_problem}
\end{equation}

where $\mathcal{F}_{n}^{*}$ denotes an optimal size-$n$ feature set w.r.t. an error function $\mathcal{L}$, $f$ is a regression model with its trainable parameters $\theta$ and hyperparameters $\lambda$. The outer optimization searches over all size-$n$ feature subsets, while the inner optimization yields the best-trained model for a specific subset.

Since the exact optimization is computationally infeasible due to the combinatorial size of $\mathcal{S}_n$ and the hyperparameter search space $\Lambda$, we instead aim to find an approximate solution with feature selection methods $\mathrm{FS}$ that are defined as a mapping $\mathrm{FS}: (\bm{X},\bm{Y}) \longrightarrow \mathcal{S}_n$ providing an approximation $\widehat{\mathcal{F}}_n := \mathrm{FS}(\bm{X},\bm{Y}) \approx \mathcal{F}^*_n.$ In the next section, we will introduce our proposed \ac{FS} method \acf{CSFS}, a novel wrapper-based \ac{FS} method aiming to improve on the computational complexity issue of wrapper-based approaches.

\subsection{Cluster-based Sequential Feature Selection}
\label{sec:approach:csfs}

Our method builds on the wrapper-based \ac{SFS} algorithm~\cite{ferri_sfs_1994}, a greedy \ac{FS} procedure that can be run either forward or backward. The backward variant starts with all features and iteratively removes the least informative one, whereas the forward variant starts empty and adds the most informative feature at each step. Since the two directions often yield different results, we use the backward mode, as we assume it might better capture joint feature utility. For example, in \ac{PV} power prediction, the \ac{GHI} and \ac{DNI} may appear weak individually, leading forward \ac{SFS} to overlook them, yet together they might enable to better estimate the effective irradiance reaching a \ac{PV} module. We presume backward \ac{SFS} is more likely to retain such complementary features. Thus, throughout this work, \ac{SFS} refers to the backward variant.


Compared to other \ac{FS} algorithms, \ac{SFS} offers multiple advantages. In general, wrapper-based methods are capable of providing a good approximation of the optimal feature set~\cite{DBLP:journals/apin/DhalA22}. Compared to \ac{RFE}, it does not rely on a \ac{FI} function, which would introduce greater complexity and dependence on the \ac{FI} method selection. An \ac{FI} function whose ranking does not reflect the actual utility for the used model could lead to inferior results, for instance, due to a bias toward high-cardinality features as for the intrinsic \ac{RF} \ac{FI}. In \ac{SFS}, selection is based solely on the model's predictive performance and is therefore optimized towards it.

However, \ac{SFS} in its original form also has considerable drawbacks. First, the number of models that need to be trained to select $n$ out of $N$ features can become considerably high, namely $0.5 (N - n)(1+N+n)$, leading to high computational costs. Second, it is prone to redundant computations because it iterates over all remaining features, even after an uninformative feature has been found, resulting in unnecessary computation time. Third, redundant computations might arise from correlated features that provide similar information but are evaluated separately.

\begin{figure}[htb]
    \centering
    \includegraphics[width=0.95\linewidth]{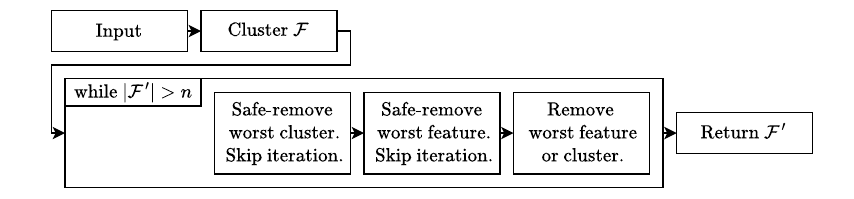}
    \caption{CSFS process starting with input data, clustering of the feature set $\mathcal{F}$, sequential safe-removal of worst cluster or feature, conditional force-remove of worst feature or cluster as fallback, returning the final feature set $\mathcal{F'}$.}
    \label{fig:csfs-approach}
\end{figure}

To mitigate these issues, we propose \ac{CSFS} (see Fig.~\ref{fig:csfs-approach}), which introduces several improvements to enable model-agnostic and computationally efficient \ac{FS}, thereby contributing to the engineering of data-driven information systems:
\begin{itemize}
    \item Instead of individual features, \ac{CSFS} operates on clusters of features. Clustering can be performed based on feature correlations, RF~FI as an initial best guess, or randomly. This allows multiple features to be removed in a single iteration. We define a valid feature clustering of feature set $\mathcal{F}$ as $\mathcal{C} = \{ c_i \mid c_i \subseteq \mathcal{F} \land \cup_i c_i = \mathcal{F} \land \cap_i c_i = \emptyset \}$.

    \item The \ac{FS} algorithm is enhanced by a non-inferiority test that allows for safe and early discarding of feature clusters if the removal does not negatively impact the predictive performance with a high certainty.

    \item If at the cluster-level no safe-removable cluster is found, a fine-grained feature-level analysis, also with non-inferiority testing, is conducted. Therefore, the higher cost of individual feature testing like in \ac{SFS} is only incurred if needed.

    \item In case no feature is safe-removable, a conditional force-remove of the worst feature or cluster is triggered.
\end{itemize}

\vspace{5cm}
\section{Experimental Setup}
\label{sec:experimental_setup}


\subsection{Software and Hardware}
\label{sec:experimental_setup:software_and_hardware}

The experiments are conducted on a server featuring an AMD EPYC 9354P 32-Core processor @3.25 GHz and 377GB of RAM. The server runs Ubuntu 24.04.4~(LTS). Experiments are executed in parallel with the number of used CPU~cores limited to two, respectively. The project code is written in Python and publicly available\footnote{https://github.com/dan-grm/CSFS-preprint} to foster reproducibility.

\subsection{Wind Power Dataset}
\label{sec:experimental_setup:wind_dataset}

We use a publicly available \ac{WT} \ac{SCADA} dataset published by \acf{EDP}~\cite{edp}. It contains two years of \ac{SCADA} records from five offshore Vestas 2~MW wind turbines in the West African Gulf of Guinea, including system-state variables such as operating temperatures, rotor and generator rotation speeds, and adjustment angles. In addition, meteorological data from a meteorological mast are provided, including wind speed and direction, humidity, and atmospheric pressure. There are two anemometer sensors at 80~m and 77~m, weather vanes at 77~m and 40~m, and temperature and pressure sensors at 75~m and 100~m height, respectively~\cite{menezes_wind_2020}. Lastly, possibly relevant datetime features are extracted from the timestamps, namely the day of the year and the time of the day. The dataset is provided at a 10-minute periodicity.


We differentiate between two dataset variants WT-S1 and WT-S2 with 46 and 15 features, respectively. In the first scenario (S1), we use all available legitimate variables, that is, system-state, meteorological, and datetime features. The goal is to obtain an accurate digital twin model that may serve as a reference for system operators to detect anomalies, as deviations between predicted and actual power may indicate sensor faults or impending component failures. In scenario S2, we only consider features such as datetime and meteorological variables that would realistically be available at forecast time, for instance, by querying \ac{NWP} models.

For our study, we select wind turbine \emph{T11} with the highest data availability, namely $87004$ samples. To split the data into training, validation, and test data, we apply a (50\%/25\%/25\%) temporal split to avoid temporal leakage~\cite{DBLP:conf/wosp/GrillmeyerHLBL025}. We use the full 2-year period of data available from 2016 and 2017; thus, we split at the timestamps 2016/12/31 3:50 a.m. and 2017/07/02 5:50 a.m. with a symmetrically inserted gap of 1~day, respectively.

\subsection{PV Power Dataset}
\label{sec:experimental_setup:pv_dataset}
We use the publicly available \acf{PVOD}, which provides metadata and environmental variables from 10 different \ac{PV} sites in Hebei Province, China~\cite{pvod}. It contains 300~days of data from 2018/07/01 to 2019/06/13 at a 15-minute periodicity. The metadata includes site-specific information, such as total installed capacity, panel size and tilt, and longitude and latitude. The meteorological and atmospheric variables include solar irradiation, ambient temperature, humidity, atmospheric pressure, and wind speed and direction. Each of these variables is available from two different sources. The first is a \ac{NWP} model, namely the Advanced Research WRF (ARW) Version 3.9.1 modeling system~\cite{yao_photovoltaic_2021,michalakes}. The second source is measurements conducted at the specific sites, denoted as \ac{LMD}.

Similar to Section~\ref{sec:experimental_setup:wind_dataset}, we differentiate between two dataset variants, PV-S1 and PV-S2, with 37 and 20 variables, respectively. In the first scenario (S1), both \ac{LMD} and \ac{NWP} variables are available. In the second scenario (S2), we assume that \ac{LMD} is unavailable, for instance, due to a lack of suitable local monitoring infrastructure, and rely solely on \ac{NWP} variables. In both cases, we assume the availability of longitude and latitude to enable the calculation of additional variables derived from solar geometry~(see Section~\ref{sec:rw:pv}).

For our study, we select \emph{station 01} with the highest data availability, that is, $33.408$~samples. The total installed capacity is 20~MW. We perform a leakage-aware temporal split with train/val/test ratios (50\%/25\%/25\%) and symmetrical 1-day gaps, resulting in the split timestamps 2018/12/21 5:00~a.m and 2019/03/19 10:00~p.m.

\subsection{Data Preprocessing}
\label{sec:experimental_setup:preprocessing}
Before using the data from the two datasets to evaluate \ac{FS} techniques and as inputs to the adopted regression models, we apply a preprocessing pipeline to convert the data into a form suitable for the models and to avoid unreliable results due to data leakage~\cite{kapoor2023leakage}. For both datasets, first, duplicates are detected and removed. Second, features with a missing value rate exceeding 5\% are removed. Third, observations with at least one missing value are removed. Fourth, features with negligible variance are discarded, as they do not carry any useful information. Fifth, cyclical encodings for circular features, for instance, the day of the year, are added for \ac{ANN} models.

For the \ac{EDP} Dataset, we identify and remove \ac{SCADA} variables that might allow for a reconstruction of the power output using direct mathematical relationships, for instance, the voltage, current, or $\cos{\phi}$, capacitive and inductive power, and estimated wind speed, as these features might thereby hinder the models to learn a meaningful system state representation in the form of a digital twin.
For the \ac{PVOD}, no proxy variables for the target variable have been identified. In \ac{PV} power generation, it is trivial that between sunset and sunrise, the power output must be zero. We remove those observations, making the dataset more balanced and facilitating model optimization to capture the non-trivial relationship between input variables and power generation during the daytime.
\section{Results}
\label{sec:results}

\subsection{Predictive Performance}
\label{sec:results:predictive_performance}

In this section, we compare the predictive performance achieved with feature sets selected by \ac{CSFS}, \ac{SFS}, and filter-based and embedded baseline methods. In each experiment, we set a fixed target feature size $|\mathcal{F}'| \in \{ 2,3,5,8,10 \}$ to ensure a fair comparison between the assessed \ac{FS} methods. We use a diverse set of \ac{ML} and \ac{DL} regression models, namely \ac{MLP}, LightGBM, and XGBoost, in conjunction with hyperparameter optimization using \ac{CFO}~\cite{wu2021frugal}. We optimize the hyperparameters once at the beginning of each \ac{SFS} or \ac{CSFS} iteration on the validation set, and for the final test of the determined feature selections of all \ac{FS} methods on the training and validation data. We adopt the most widely used \ac{RMSE} as the primary metric, which penalizes large deviations, measures errors in the data's original units, and remains stable for values near zero~\cite{DBLP:conf/wosp/GrillmeyerJHLK26}.

\begin{figure}[h]
    \centering
    \includegraphics[width=\linewidth]{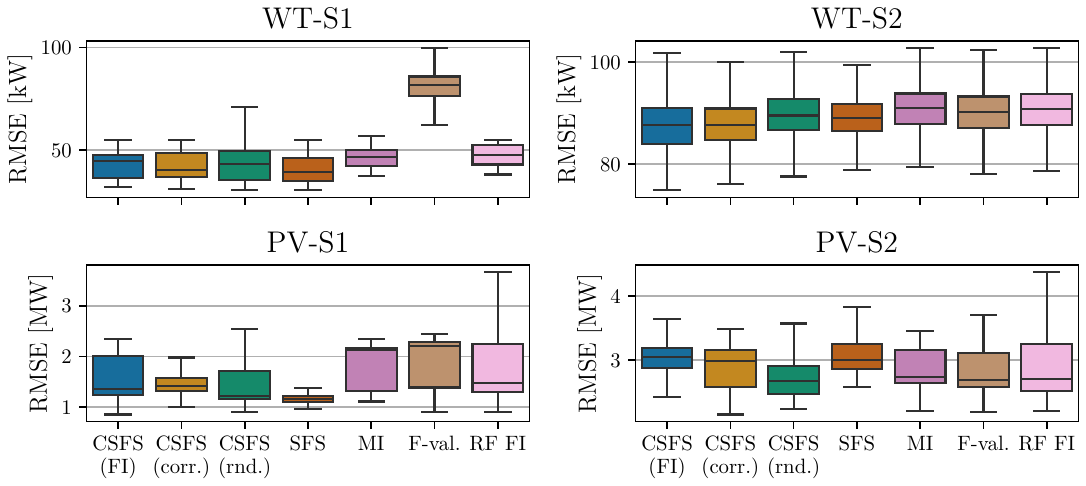}
    \caption{Predictive performance of all tested \ac{FS} techniques, aggregated over all models and target feature sizes for the four datasets WT-S1, WT-S2, PV-S1 and PV-S2.}
    \label{fig:eval:agg_performance}
\end{figure}

To evaluate the overall test performance of the feature sets obtained by different FS methods, we first analyze aggregated results across all models and feature set sizes. Fig.~\ref{fig:eval:agg_performance} shows the aggregated \ac{RMSE} for all datasets. The x-axis indicates the \ac{FS} techniques: \ac{CSFS} with three clustering methods, \ac{SFS}, the filter-based methods F-value and \ac{MI}, and the embedded \ac{RF} \ac{FI} approach. For a fair comparison between CSFS and SFS, we employ an enhanced SFS version that also uses the safe-remove shortcut at the feature level whenever non-inferiority is provided. Despite the high IQR of the results due to clear differences in feature set sizes, they enable an overall global assessment.

Fig.~\ref{fig:eval:agg_performance} reveals that for WT-S1, WT-S2, and PV-S1, the wrapper-based methods \ac{SFS} and \ac{CSFS} show consistently better median results than the simpler baseline methods. However, for WT-S2, the differences in \ac{RMSE} are relatively small and amount to only a few kilowatts. For PV-S2, the baseline methods \ac{MI}, F-value , and \ac{RF} \ac{FI} perform surprisingly well with an \ac{RMSE} of about 2.7~MW. Only the \ac{CSFS} method based on random grouping performs slightly better. Overall, the results suggest that wrapper-based methods tend to provide an advantage primarily for the larger dataset variants. Among the wrapper-based methods, none of the three \ac{CSFS} variants appears superior, and their ranking is dataset-dependent. Most notably, overall \ac{CSFS} performs on par with \ac{SFS}, which exhibits slightly better median performance and lower variability for PV-S1.

\begin{figure}[h]
    \centering
    \includegraphics[width=\linewidth]{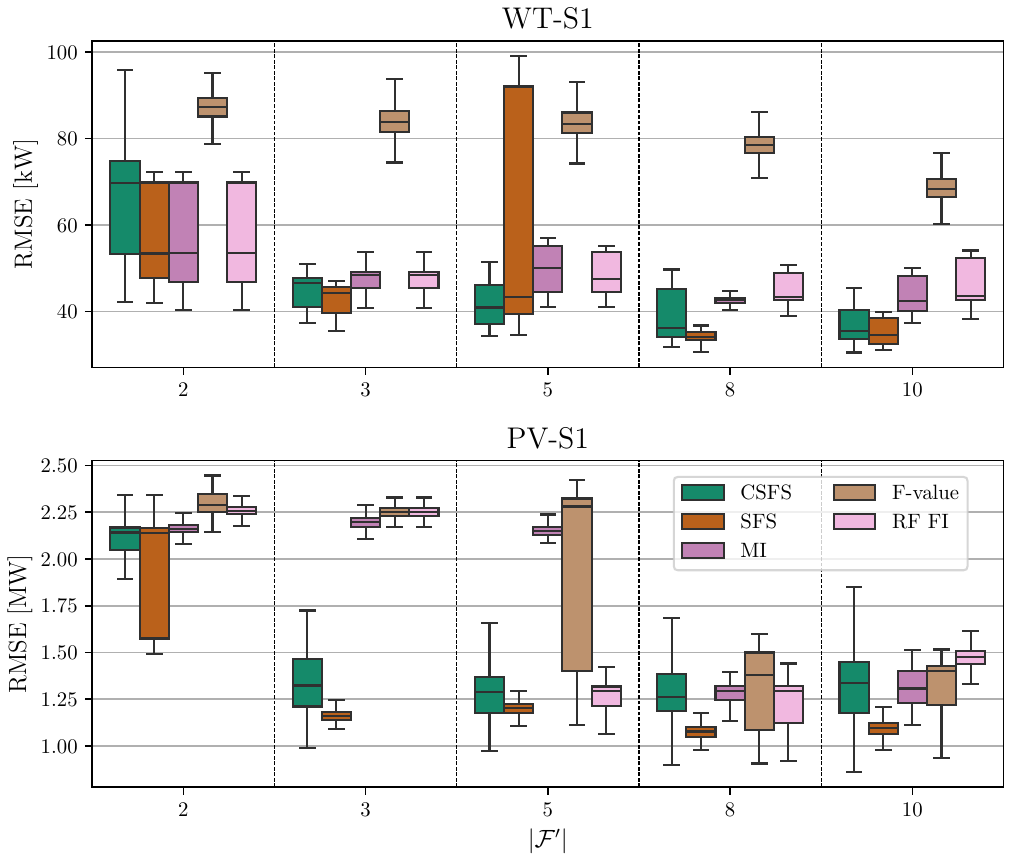}
    \caption{Predictive performance for different target feature set sizes and \ac{FS} methods on the WT-S1 and PV-S1 datasets.}
    \label{fig:eval:rmse_over_n}
\end{figure}

Further, Fig.~\ref{fig:eval:agg_performance} shows that the predictive performance achievable with dataset variant S1 is considerably better than with variant S2 for both domains (wind and PV): While for WT-S1, most methods achieve a median \ac{RMSE} below 50~kW, the median \ac{RMSE} for WT-S2 is around 90~kW. Similarly, while the achieved \ac{RMSE} for PV-S1 mostly lies between 1~MW and 2~MW, the \ac{RMSE} for PV-S2 increases to approximately 2.7~MW to 3~MW. These differences can be attributed to the additional system and local measurement variables in variant S1, from which the models can evidently derive benefit.

In the following, we further analyze how the target feature size affects the predictive performance of the complete-dataset variants WT-S1 and PV-S1. Therefore, Fig.~\ref{fig:eval:rmse_over_n} shows the performance broken down by dataset variants and feature-set sizes, and aggregated across the three models used. For CSFS, the values are additionally aggregated across the different clustering methods.

The upper plot shows the performance on the WT-S1 dataset. It reveals that for two features, SFS, \ac{MI} and RF~FI perform similarly with median \ac{RMSE} values of about 55~kW, while the other methods perform considerably worse. The aggregated result of \ac{CSFS} appears particularly poor in this case because the variants using FI-based and random clustering yield RMSE values of 77~kW and 80~kW, respectively, whereas the correlation-based clustering performs substantially better with an RMSE of approximately 46~kW.
For larger feature sets, the wrapper-based methods are consistently better than the simpler baselines, reaching a median \ac{RMSE} of about 35~kW, while the other methods reach at most 43~kW. Except for two features, \ac{CSFS} performs only slightly worse than \ac{SFS}, and for five features even slightly better and more stable. Overall, the F-value-based \ac{FS} clearly performs the worst.

For the PV-S1 dataset in Fig.~\ref{fig:eval:rmse_over_n}~(lower), the wrapper-based methods \ac{SFS} and \ac{CSFS} show consistent improvements from two up to eight features, where they reach their optimal median \ac{RMSE} of 1.08~MW and 1.25~MW, respectively. The simpler baseline \ac{FS} methods also improve with larger target sizes, but not to the same extent, achieving a median \ac{RMSE} of at most 1.29~MW for eight features. As in WT-S1, the F-value-based \ac{FS} yields the least stable results.

\subsection{Runtime Comparison}
\label{sec:results:runtime_comparison}
\begin{figure}[h]
    \centering
    \includegraphics[width=1\linewidth]{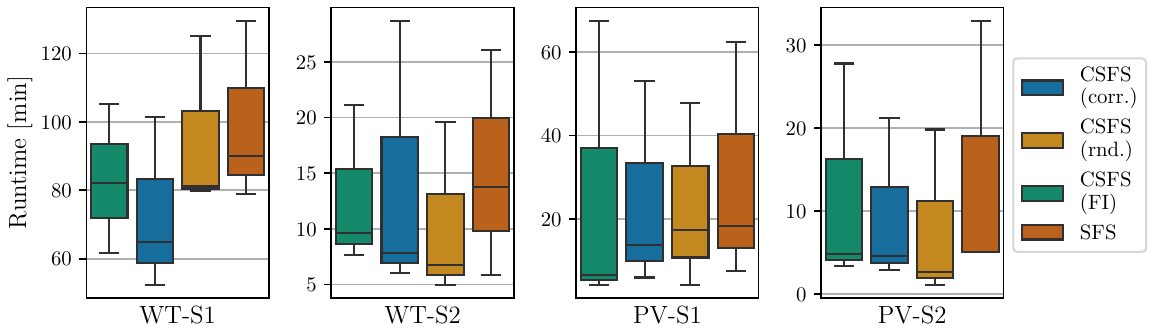}
    \caption{Runtime comparison of \ac{SFS} and \ac{CSFS} with three different clustering methods across four dataset variants. The values are aggregated over all used models for the target set feature size $|\mathcal{F'}|=2$.}
    \label{fig:runtime_comparison}
\end{figure}

The drawback of wrapper-based methods is their computational complexity, while the runtime of filter-based methods is comparably negligible, as explained in Section~\ref{sec:rw:feature-selection}. Therefore, we focus on the runtime comparison of the wrapper-based methods \ac{CSFS} and \ac{SFS}. In Fig.~\ref{fig:runtime_comparison}, we visualize the aggregated runtime of \ac{SFS} and \ac{CSFS} for three distinct clustering techniques across all dataset variants.

The median runtime of all methods strongly depends on the dataset scenario and, therefore, the initial feature set size. For WT-S1 and PV-S1, \ac{SFS} shows median runtimes of about 90~min and 18~min, respectively, with large variations. For the smaller datasets WT-S2 and PV-S2, the \ac{SFS} runtime is greatly reduced to 14~min and 5~min, respectively.

The median runtime of all \ac{CSFS} variants is consistently lower, leading to an overall runtime reduction of $19.4\%$, $9.3\%$, $25.3\%$, and $39.8\%$ in comparison to \ac{SFS} for the four dataset scenarios, respectively.

Notably, in our experiments, the \ac{SFS} baseline includes the same safe-remove shortcut at the feature level as used in \ac{CSFS} (see Section~\ref{sec:approach:csfs}), making it more computationally efficient than in its original form. Therefore, the reported runtime improvements are conservative.

\subsection{Domain-Level Analysis}
\label{sec:results:domain_level_analysis}
To explore domain-specific insights, Tab.~\ref{tab:top_10_features} presents the ten most important features per dataset for the best-performing feature set and model found across all experiments. For both energy domains, scenario S1 shows considerably better results in all performance metrics. This indicates that the models can clearly benefit from system-level and local measurement data, respectively. The best feature subsets of size ten for the dataset variants WT-S1, WT-S2, PV-S1, and PV-S2, have been found by SFS, CSFS, CSFS, and F-value, respectively.

\begin{table}[]
\caption{Top ten features identified for each dataset and scenario. Features selected in both scenarios are highlighted in \textbf{bold} for each dataset.}
\begin{threeparttable}
\adjustbox{max width=\textwidth}{
\begin{tabular}{l@{\hskip 0.1in}l@{\hskip 0.1in}l@{\hskip 0.1in}l@{\hskip 0.1in}l}
\toprule
                                   & WT-S1                                   & WT-S2                         & PV-S1                         & PV-S2                         \\ \midrule
\multicolumn{1}{c}{\hspace{-0.4cm}\multirow{10}{*}{\begin{sideways}Selected features\end{sideways}}} & \textbf{Amb. Temp.}            & Abs. Wind Dir.       & LMD App. Zenith\tnote{b}         & \textbf{MinuteOfDay\tnote{b}}    \\
                                    & \textbf{Amb. WS}             & \textbf{Amb. Temp.}  & LMD Diff. Irrad.              & \textbf{NWP App. El.\tnote{b}} \\
                                    & Blade Pitch Angle                       & Amb. Temp. 2         & LMD GHI                       & NWP App. Zenith\tnote{b}         \\
                                    & Gear Bear. Temp.                & \textbf{Amb. WS}   & LMD Glob. Irrad.              & NWP Azimuth\tnote{b}             \\
                                    & Generator RPM                           & MinuteOfDay                   & \textbf{MinuteOfDay\tnote{b}}    & NWP DNI                       \\
                                    & Grd. R. Inv. Ph.2 T. & Precipitation                 & \textbf{NWP App. El.\tnote{b}} & \textbf{NWP Dir. Irrad.}      \\
                                    & Nacelle Dir.                       & \textbf{Turb. Int.} & NWP App. El.\tnote{a}          & NWP El.\tnote{b}               \\
                                    & Nacelle Temp.                     & Turb. Int.        & \textbf{NWP Dir. Irrad.}      & NWP GHI                       \\
                                    & \textbf{Turb. Int.}           & Turb. Int. 2        & \textbf{NWP Glob. Irrad.}     & \textbf{NWP Glob. Irrad.}     \\
                                    & VCP Water Temp.                   & Wind Speed 1                  & \textbf{NWP Zen.\tnote{b}}     & \textbf{NWP Zen.\tnote{b}}     \\
                                    \midrule
FS                                  & SFS                                     & CSFS (FI)                     & CSFS (FI)                     & F-value                       \\
Model                               & XGBoost                                 & XGBoost                       & MLP                           & MLP                           \\
\midrule
RMSE                                & $(32.49 \pm 0.30)$ kW                     & $(82.4 \pm 3.6)$ kW             & $(0.971 \pm 0.047)$ MW          & $(2.40 \pm 0.10)$ MW            \\
MAE                                 & $(20.64 \pm 0.16)$ kW                     & $(26.90 \pm 0.48)$ kW           & $(0.615 \pm 0.028)$ MW          & $(1.593 \pm 0.091)$ MW          \\
R2 [\%]                                  & $99.6632 \pm 0.0069$                 & $97.83 \pm 0.19$           & $97.49 \pm 0.25$           & $84.6 \pm 1.3$             \\
\bottomrule
\end{tabular}
}
\vspace{0.03cm}
\begin{tablenotes}
\item[a/b] Sine/Cosine transform. applied for MLP \scriptsize{(Note: Tree-based models do not require it)}.
\end{tablenotes}
\end{threeparttable}
\label{tab:top_10_features}
\end{table}

For the \ac{WT} dataset, both scenarios incorporate ambient wind speed (WS) and temperature, and turbulence intensity data. While for WT-S2 additional environmental measurements like wind direction, precipitation, and time information are used, WT-S1 rather exploits system-state variables like the blade pitch angle, generator RPM, and gear bearings, spinner, grid rotor inverter, and VCP temperatures. This allows for more accurate modeling of the power production with absolute improvements of ($49.95\pm3.58$)~kW and ($6.26\pm0.51$)~kW in terms of \ac{RMSE} and \ac{MAE}, respectively.

In \ac{PV} power prediction, both scenarios incorporate time information and the \ac{NWP} variables apparent elevation, direct and global irradiance, and the zenith. In PV-S1, the model benefits from additional \ac{LMD} variables, namely the apparent zenith, diffuse and global irradiance, and \ac{GHI}, which allows for considerable absolute \ac{RMSE} and \ac{MAE} improvements, namely ($1.43\pm0.11$)~MW and ($0.979\pm0.095$)~MW, respectively.


\subsection{Discussion}
\label{sec:results:discussion}

The analysis has shown that, in summary, wrapper-based \ac{FS} methods overall yield better-performing feature selections than simple F-value- or \ac{MI}- based methods or the \ac{RF} \ac{FI}, confirming findings from prior work~\cite{DBLP:journals/apin/DhalA22}. This comes at the cost of a higher computational complexity.

Our proposed approach, \ac{CSFS}, achieves performance only slightly worse than or on par with the enhanced \ac{SFS} version with safe-remove shortcuts applied, while running substantially faster, on average by 20.8\% across all experiments. This can be explained by the more efficient early removal of whole feature clusters in 29.5\% of all \ac{CSFS} iterations, on average. The domain-level analysis confirms the assumption of other authors~\cite{pvod,machines9050100} that both system-level and \ac{LMD} variables can improve accuracy.


Potential threats to validity include distortions resulting from the limited dataset, particularly the use of only a single \ac{WT} and \ac{PV} station. Moreover, employing a different primary evaluation metric instead of \ac{RMSE}, or adopting alternative hyperparameter search spaces, could lead to different outcomes. Finally, repeating the \ac{FS} experiments with different random seeds may alter the decisions made in individual iterations. Due to the greedy nature of the algorithm, such variations can propagate throughout the procedure and ultimately result in different feature sets being selected. At the same time, measures were taken to account for statistical variability: the evaluations were complemented with test-set bootstrapping, and the final test evaluation was repeated ten times to obtain more robust performance estimates.
\section{Conclusion}
\label{sec:conclusion}
The growing awareness of the impacts of climate change, combined with the globally increasing energy demand, automatically leads to greater use of renewable energy. 
While conventional energy suppliers are able to control the demanded energy output reliably, power produced by renewable energies highly fluctuates, driven by environmental factors.
Existing literature, analyzed in our structured literature review, often relies on a small subset of available features, which lets us assume unexploited potential in remaining features.
Therefore, we propose a novel model-agnostic cluster-based methodology for feature selection and provide a scikit-learn compatible open source implementation.
In our evaluation, we show that the wrapper-based techniques overall provide better-performing selections of features than simpler filter-based or embedded baseline methods. Thereby, our proposed approach CSFS achieves a predictive performance comparable to SFS while reducing computational cost by an average of 21\%.
Hence, we consider the presented approach a reasonable alternative, providing faster time-to-result while delivering reasonable results.

In the future, we would like to delve even deeper into the evaluation and apply CSFS with further clustering techniques and in combination with Transformer-based time series forecasting models.
Furthermore, we will test and evaluate the approach in additional domains, including those outside the renewable energy sector.
Regarding the proposed methodology, we would like to develop a hybrid variant that, in addition to the existing backward feature selection method, also performs forward feature selection in parallel, and test possible extensions, e.g., through additional dimensionality reduction with PCA.

\begin{credits}

\subsubsection{\discintname}
The authors have no competing interests to declare that are relevant to the content of this article.
\end{credits}
%
%
%
\bibliographystyle{splncs04}
\bibliography{base}
%




\end{document}